\documentclass[conference]{IEEEtran}
\IEEEoverridecommandlockouts
\usepackage{cite}
\usepackage{amsmath,amssymb,amsfonts}
\usepackage{algorithmic}
\usepackage{graphicx}
\usepackage{subcaption}
\usepackage{textcomp}
\usepackage{xcolor}
\usepackage{orcidlink}
\usepackage{comment}
\usepackage{multirow}
\usepackage{gensymb}
\usepackage{siunitx}

\usepackage{hyperref}

\hypersetup{
  colorlinks=false,
  urlcolor=black,
  linkcolor=black,
  linkbordercolor=white,
 urlbordercolor=white,
pdfborder={0 0 0}
} 

\def\BibTeX{{\rm B\kern-.05em{\sc i\kern-.025em b}\kern-.08em
    T\kern-.1667em\lower.7ex\hbox{E}\kern-.125emX}}
\begin{document}

\title{Urban Air Pollution Forecasting: a Machine Learning Approach leveraging Satellite Observations and Meteorological Forecasts  \\
\thanks{This work was developed in the context of the Horizon Europe project UP2030 (grant agreement n.101096405).}
}

\author{
    \IEEEauthorblockN{
        Giacomo Blanco \orcidlink{0000-0001-6500-8511}, 
        Luca Barco \orcidlink{0000-0002-9089-9616}, 
        Lorenzo Innocenti \orcidlink{0009-0009-6280-6230}, 
        and
        Claudio Rossi \orcidlink{0000-0001-5038-3597}}
        \\
    \IEEEauthorblockA{
                    LINKS Foundation\\ Turin, Italy \\
                    \textit{\{name\}.\{surname\}@linksfoundation.com}
    }
}

\maketitle

\begin{abstract}
Air pollution poses a significant threat to public health and well-being, particularly in urban areas. This study introduces a series of machine-learning models that integrate data from the Sentinel-5P satellite, meteorological conditions, and topological characteristics to forecast future levels of five major pollutants. The investigation delineates the process of data collection, detailing the combination of diverse data sources utilized in the study. Through experiments conducted in the Milan metropolitan area, the models demonstrate their efficacy in predicting pollutant levels for the forthcoming day, achieving a percentage error of around 30\%. The proposed models are advantageous as they are independent of monitoring stations, facilitating their use in areas without existing infrastructure. Additionally, we have released the collected dataset to the public, aiming to stimulate further research in this field. This research contributes to advancing our understanding of urban air quality dynamics and emphasizes the importance of amalgamating satellite, meteorological, and topographical data to develop robust pollution forecasting models.
\end{abstract}


\section{Introduction}
Air pollution remains a critical global issue, affecting both human health and the environment. Across cities and regions, harmful pollutants such as PM2.5, ozone, and nitrogen dioxide pose significant risks to respiratory and cardiovascular health. As urbanization and industrialization continue to grow, addressing air quality becomes paramount. According to a study by the European Environmental Agency, in 2021, $97\%$ of the urban population was exposed to concentrations of fine particulate matter above the health-based guideline level set by the World Health Organization \cite{EEA2023}. Among these urban areas, Milan, is no exception, being among the Italian cities that have registered the highest number of days with a pollution level over the recommended thresholds in 2023 \cite{LegAmb}.

In this context, the implementation of a predictive tool for air quality, capable of forecasting pollution levels in the coming days, proves important for both citizens and decision-makers. 
For citizens, such a tool offers proactive insights into air quality, empowering them to adopt informed behaviours, like limiting outdoor activities during periods of heightened pollution. At the same time, this tool serves policymakers as a support in defining targeted policies for the management and mitigation of air pollution.
With recent advancements in remote sensing technologies, satellite data has emerged as a promising and cost-effective means for monitoring and forecasting tasks. Leveraging satellite imagery offers a unique opportunity to acquire continuous measurements of air pollutant levels across vast areas. Employing machine learning algorithms facilitates the generation of forecasts based on these observations. This study focuses on the development of a suite of machine-learning models trained using pollutant levels recorded by monitoring stations, along with multi-temporal satellite data integrated with weather and topographical information for pollutant forecasting. Furthermore, we have made the collected dataset publicly available to encourage further research in this domain\footnote{\url{https://huggingface.co/datasets/links-ads/mil-qualair}}. The proposed models offer the advantage of not directly relying on monitoring stations, thus enabling their applicability even in areas lacking such infrastructure. To evaluate the performance of the proposed models, we present results obtained through cross-validation conducted annually, yielding a percentage error of approximately $30\%$ for each pollutant. In summary, our contributions encompass the demonstration of the viability of machine learning models for pollutant forecasting and the dissemination of the collected dataset to stimulate additional research in this area.

\section{Related works}
\label{sec:related}
The importance of accurate air quality forecasting has grown significantly, driven by its health implications and environmental impact. This task is complex due to the limited data availability and to the dynamic nature of the atmosphere and the myriad of factors that influence air quality, including meteorological conditions, topographical features, human activities, and chemical transformations in the atmosphere. For these reasons, several works propose different approaches to face these challenges. Furthermore, recent advancements in satellite technology and data science have opened new avenues for enhancing the accuracy of air quality forecasting.

\subsection{Air quality forecasting techniques}
Air quality forecasting traditionally relies on ground/based monitoring stations and statistical models. Zhang et al. \cite{ZHANG2022119347} provide a comprehensive review of existing methods for this task. They distinguish between deep learning and non-deep learning methods, the latter is split into two categories: \textit{Deterministic methods} and \textit{Statistical methods}. Deterministic models include chemical transport models (CTM) such as Community Multiscale Air Quality (CMAQ) and Nested Air Quality Prediction Model System (NAQPMS). CMAQ is a modelling system that simulates the emission, transport, transformation and deposition of air pollutants developed by the U.S. Environmental Protection Agency (EPA)  \cite{cmaq2}. NAQPMS is an air quality prediction model developed in China that simulates the dispersion and chemical transformation of air pollutants \cite{ NAQPMS2}. 
Statistical models include traditional machine learning algorithms such as Autoregressive–Integrated Moving Average (ARIMA), a statistical analysis model that aims to describe time series data and forecast future points in the series \cite{arima1}. 

\subsection{Data in air quality monitoring}
The interplay between air quality and meteorological conditions is well-documented, with weather variables significantly influencing pollutant dispersion and concentration levels, as shown in several works. Tateo et al. \cite{atmos14030475} present a case study to explore the association between meteorological conditions and air quality in Lecce, a city in South Italy. Arnaudo et al. \cite{app10134587} present another case study in the city of Milan, in North Italy including also vehicular traffic.  Furthermore, Méndez et al.\cite{MendezML} show that the combination of pollutant features and weather features provides the best accuracy.
In recent years,  Reshi et al. \cite{S5-app} and Can Li et al. \cite{LI20113663} demonstrate the importance of adding satellite data, i.e. Sentinel-5P data, to obtain better results.

\section{Dataset}
\label{sec:dataset}
\begin{figure}[bht]
\centering
\includegraphics[width=0.9\linewidth]{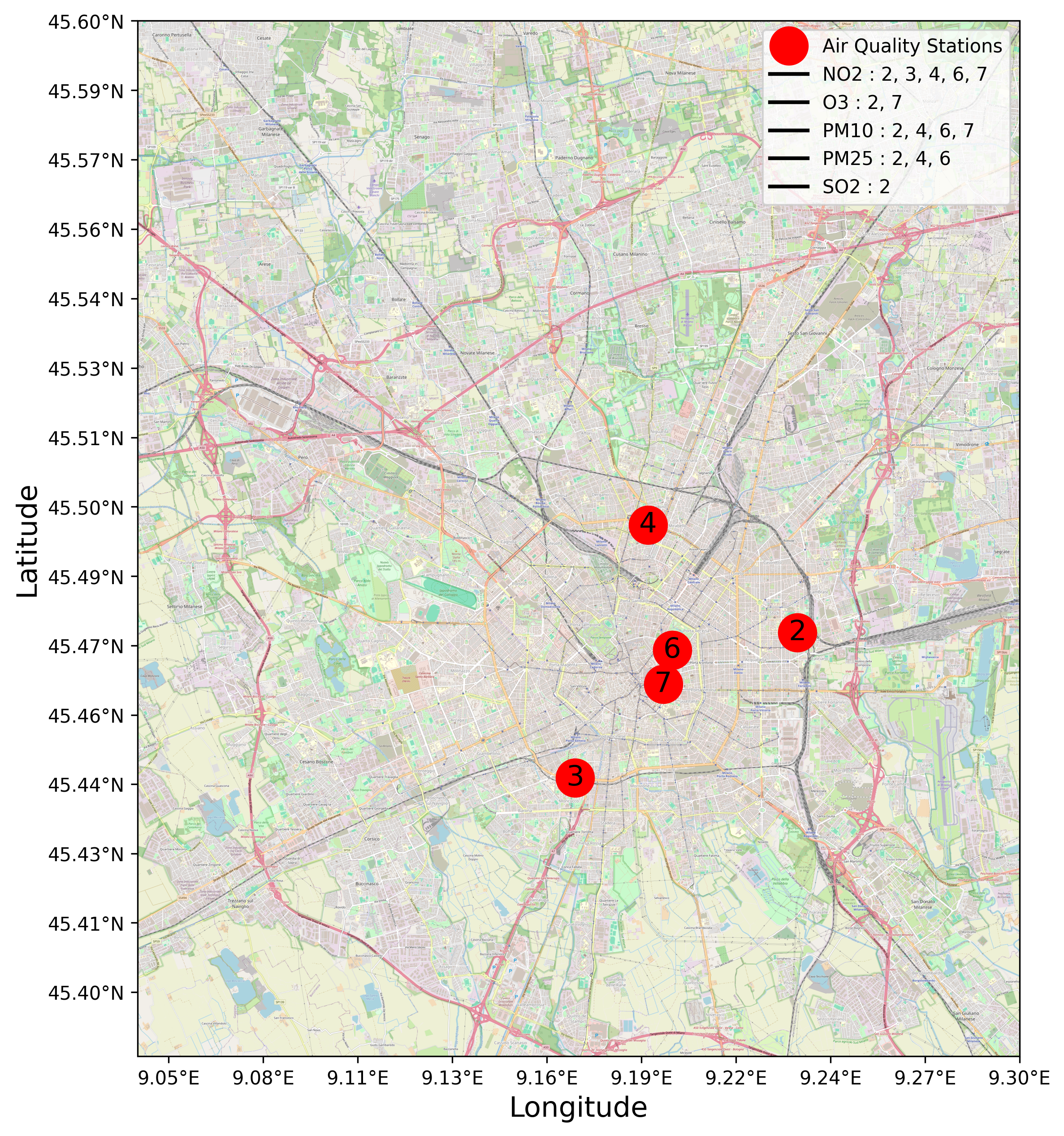}
\caption{Map of the Milan area considered in this study. Pollutant monitoring stations are marked as red dots, each labelled with its station id. The legend includes stations measuring each pollutant.}
\label{fig:stations}
\end{figure}

\begin{figure*}[bht]
\centering
\includegraphics[width=0.65\linewidth]{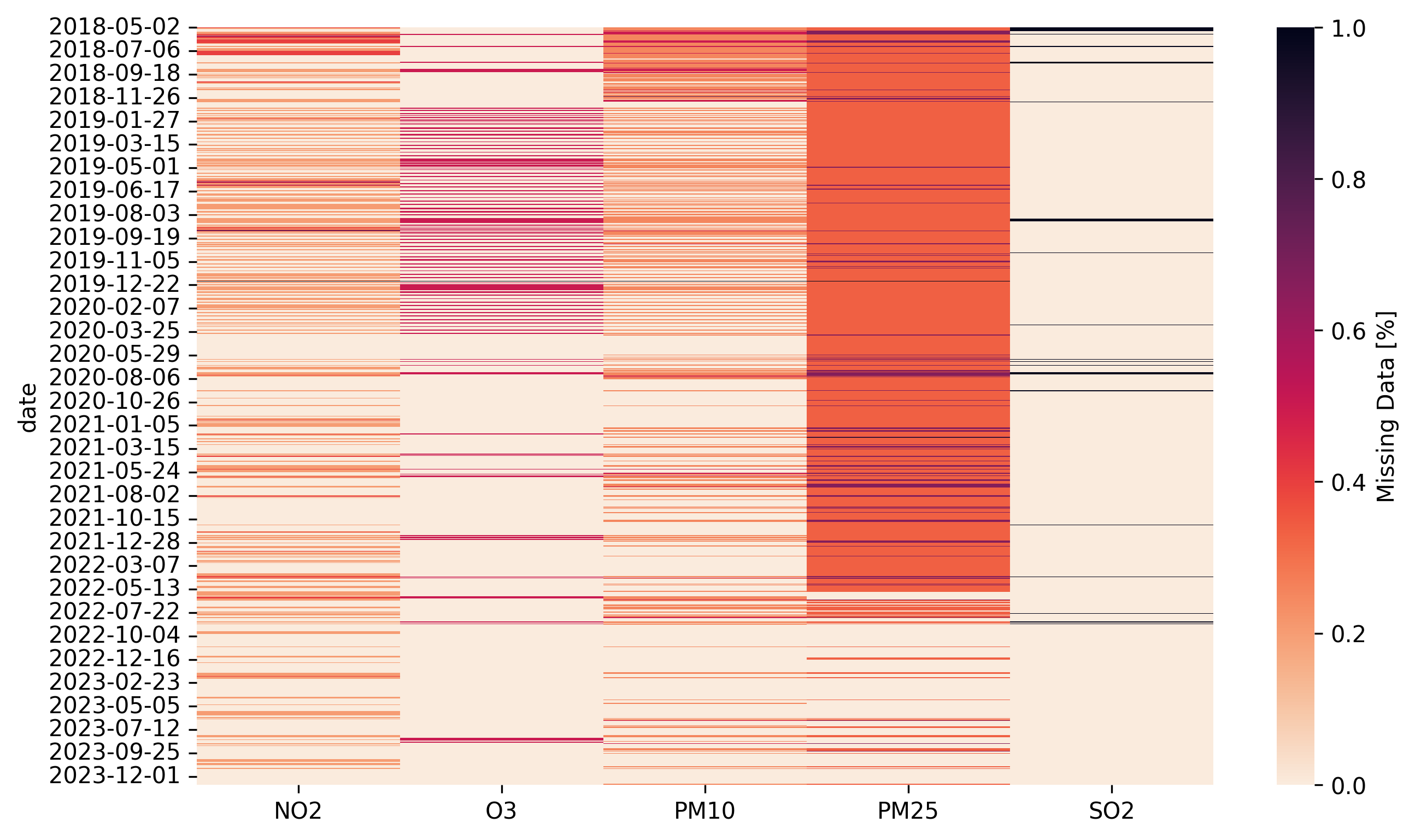}
\caption{Number of missing values for each pollutant over the reference period, normalized by the total number of monitoring stations for each pollutant.}
\label{fig:missing}
\end{figure*}

This study utilizes multiple data sources, serving as inputs to the proposed machine learning model and as ground truth references. Input sources include remote sensing data along with weather and topographical data. The latter serves as static sources, remaining unchanged over the study period, whereas remote sensing and weather data are updated daily. The following subsections delve into a detailed description of these data sources, outlining the gathering process and the preprocessing techniques applied.

\subsection{Pollution Data}
The dataset containing pollutant measurements from ground stations serves as the ground truth for our machine-learning models. This dataset has been obtained from the open data portal of the Municipality of Milan \footnote{\url{https://dati.comune.milano.it/}}. Daily measurements of various pollutant concentrations, expressed in $\mu g/m^3$, have been available since 2003. However, due to the availability of satellite data only from May 2018 onwards, we restricted the time frame from this date until the end of 2023. Among the available pollutants, our focus is on the five pollutants used to evaluate the Air Quality Index (AQI) as proposed by the European Environmental Agency: $P=\{PM10, PM2.5, NO2, SO2, O3\}$. There are a total of five monitoring stations distributed across the Milan area and, as depicted in Figure \ref{fig:stations}, not all stations measure every pollutant. In addition, there are intermittent instances of missing measurements. This discrepancy is illustrated in Figure \ref{fig:missing}, where we present the percentage of missing values for each day, computed across the stations expected to measure the respective pollutant. This dataset serves as a fundamental component for training and validating our machine learning models, facilitating predictions based on locally observed air quality conditions.

\subsection{Satellite Data}
The utilization of satellite data as an input for the proposed models aims to broaden their applicability to regions where station density is limited or even absent. The Copernicus Sentinel-5P mission, an integral component of the European Union's Copernicus program, serves this purpose by focusing on monitoring the composition of the Earth's atmosphere. Equipped with the Tropospheric Monitoring Instrument (TROPOMI), this satellite delivers daily data with a spatial resolution of $5.5 \times 3.5$ km concerning the concentration of various atmospheric pollutants, including nitrogen dioxide, ozone, sulfur dioxide, formaldehyde, carbon monoxide, and the aerosol index value indicating the presence of elevated aerosol layers in the atmosphere. As for the stations, the considered data spans from May 2018 to December 2023 and is obtained as a GeoTIFF file containing the daily measurements of the specified bands over the area depicted in Figure \ref{fig:stations}.

To ensure data consistency with station measurements, all collected files undergo the following processing. Specifically, we compute the average value of each band measured within a $500$-meter radius of each monitoring station. This approach yields five distinct measurements for each band per day, reflecting the varied local conditions surrounding the stations. Consequently, this process results in obtaining as many time series of daily satellite measurements as the number of stations.

\subsection{Weather Data}
Given the significant influence of meteorological conditions on air quality, the incorporation of this data plays a pivotal role in comprehending and forecasting variations in air pollution levels. Meteorological data, obtained from a third-party provider, encompass measurements of various atmospheric indicators, including temperature, dew point, humidity, precipitation, wind speed, wind direction, pressure, cloud cover, and solar radiation, all available at a daily resolution. Regrettably, the data lacks spatially resolved information, offering a single value for the entire area of interest. However, the acquired weather data exhibit no missing values, obviating the need for data-cleaning procedures.

While we aimed to include weather forecast data to aid in anticipating future air pollution levels, we encountered challenges in finding a provider for historical weather forecasts. To address this limitation, we utilized observed weather data on day $t$ as a proxy for the same day in the future when forecasting pollution. As a pre-processing operation, all weather variables were normalized to the range $[0,1]$, except for the wind direction variable, which is expressed as the angle in degrees between the wind and the north, ranging from $0$ to $360$ degrees. To normalize this variable and account for its circular nature, it was replaced with a trigonometric encoding represented by the pair of values obtained from the functions $\phi_{WDsin} = \sin(\pi*\alpha/180)$ and $\phi_{WDcos} = \cos(\pi*\alpha/180)$, where $\alpha$ represents the original angle in degrees.

\subsection{Topological Data}
Topological data serve as static sources in this study, providing insights into the area surrounding the monitoring stations. By integrating these data into the model, it becomes possible to discern the relationship between terrain characteristics and pollutant concentrations.

Two types of topological data are considered. Firstly, the Digital Elevation Model (DEM) offers information about terrain morphology, presenting a constant observation throughout the reference period. For each monitoring station, altitude data is extracted at three different scales: precise point altitude, altitude within a $100$-meter radius, and altitude within a $1$-kilometer radius.
Secondly, data derived from the Copernicus Urban Atlas\footnote{\url{https://land.copernicus.eu/en/products/urban-atlas}} is utilized, as it offers detailed land cover and land use maps for urban areas. Although the dataset was last updated in 2018, it provides valuable insights into land use patterns over the reference period. The original $27$ land cover classes are grouped into $10$ broader categories (including urban, road, railways, port, airports, extraction, no use, green, open spaces and water) facilitating the analysis. For each monitoring station, the percentage of area covered by each land use category within a $500$-meter radius is calculated, enabling the assessment of urbanization levels and green space presence in station surroundings.

\subsection{Data integration}
Aligning the various data sources is imperative if they will be the input for the machine learning models, despite their differing spatial and temporal resolutions. Station readings and satellite data offer information with daily temporal resolution and station-level spatial resolution. In contrast, meteorological data, while updated daily, have city-level resolution. Finally, topological data offer spatial detail at the individual station level but remain static throughout the reference period.
As the first data integration step, satellite and topological data are merged due to their shared spatial resolution. This results in the topological data being replicated for each day across the reference period. These combined datasets are then enriched with meteorological data, which share daily temporal resolution. This involves replicating the measurements for a given day across all active stations. The resulting data are aggregated daily and at the station level, facilitating integration with ground truth station data.
These aligned datasets are utilized to train machine learning models capable of predicting pollutant concentration on a given day $t$ using data from day $t-1$. To provide the model with a relevant set of features, data from the previous $w$ days are included. This results in reorganising the final dataset in $w$-day windows with a one-day step. Temporal features, such as day of the year, month of the year, and day of the week, are added to capture seasonal and weekly patterns. These features are encoded as sinusoidal functions calculated on the last day of each window. The functions are $\phi_{fsin} = sin(2\pi*f/P_f), \phi_{fcos} = cos(2\pi*f/P_f)$ where $P_f \in \{366,12,7\}$ is the period for each feature and $f \in [0,P_f]$ is the value for each feature in the last day of each window. 
The dataset associates each day $t$ with pollutant concentrations from supporting stations and satellite, meteorological, and topological data within the $[t-1, t-w]$ day interval, along with temporal features from day $t-1$ and expected weather conditions for day $t$. Possible missing values in satellite data are filled using linear interpolation during integration while missing station readings at day $t$ result in removing the corresponding record.

\begin{table*}[]
\centering
\caption{Models results for all pollutants and various temporal window values. Overall, Gradient Boosting demonstrates the best performance across all pollutants, with the exception of SO2. Results obtained with different temporal window lengths vary depending on the pollutant.}
\label{tab:results}
\resizebox{0.7\textwidth}{!}{%
\begin{tabular}{lllllllllll}
\hline
\multirow{2}{*}{\textbf{Pollutant}} & \multirow{2}{*}{\textbf{Model}} & \multicolumn{3}{c}{\textbf{w = 1}} & \multicolumn{3}{c}{\textbf{w = 7}} & \multicolumn{3}{c}{\textbf{w = 14}} \\ \cline{3-11} 
 &  & \textbf{MAE} & \textbf{MAPE} & \multicolumn{1}{l|}{\textbf{RMSE}} & \textbf{MAE} & \textbf{MAPE} & \multicolumn{1}{l|}{\textbf{RMSE}} & \textbf{MAE} & \textbf{MAPE} & \textbf{RMSE} \\ \hline
\multirow{3}{*}{\textit{PM10}} & LR & 9.562 & 0.373 & \multicolumn{1}{l|}{12.81} & 9.327 & 0.365 & \multicolumn{1}{l|}{12.4} & 9.825 & 0.390 & 12.9 \\
 & GradBst & \textbf{8.745} & \textbf{0.312} & \multicolumn{1}{l|}{\textbf{12.46}} & \textbf{8.198} & \textbf{0.294} & \multicolumn{1}{l|}{\textbf{11.52}} & \textbf{8.164} & \textbf{0.296} & \textbf{11.45} \\
 & SGDReg & 11.47 & 0.442 & \multicolumn{1}{l|}{15.62} & 10.89 & 0.423 & \multicolumn{1}{l|}{14.66} & 10.63 & 0.412 & 14.44 \\ \hline
\multirow{3}{*}{\textit{PM25}} & LR & 7.419 & 0.480 & \multicolumn{1}{l|}{10.01} & 7.325 & 0.479 & \multicolumn{1}{l|}{9.788} & 7.633 & 0.507 & 10.1 \\
 & GradBst & \textbf{6.492} & \textbf{0.372} & \multicolumn{1}{l|}{\textbf{9.546}} & \textbf{6.11} & \textbf{0.371} & \multicolumn{1}{l|}{\textbf{8.678}} & \textbf{6.138} & \textbf{0.369} & \textbf{8.728} \\
 & SGDReg & 8.624 & 0.544 & \multicolumn{1}{l|}{11.97} & 8.374 & 0.539 & \multicolumn{1}{l|}{11.44} & 8.16 & 0.526 & 11.2 \\ \hline
\multirow{3}{*}{\textit{O3}} & LR & 14.1 & 0.426 & \multicolumn{1}{l|}{17.8} & 14.64 & 0.423 & \multicolumn{1}{l|}{18.54} & 15.4 & 0.465 & 19.66 \\
 & GradBst & \textbf{12.55} & \textbf{0.338} & \multicolumn{1}{l|}{\textbf{15.94}} & \textbf{13.43} & \textbf{0.372} & \multicolumn{1}{l|}{\textbf{16.99}} & \textbf{13.69} & \textbf{0.361} & \textbf{17.47} \\
 & SGDReg & 17.98 & 0.574 & \multicolumn{1}{l|}{22.76} & 17.63 & 0.600 & \multicolumn{1}{l|}{22.17} & 17.87 & 0.557 & 22.58 \\ \hline
\multirow{3}{*}{\textit{NO2}} & LR & 15.5 & 0.265 & \multicolumn{1}{l|}{\textbf{20.21}} & 15.71 & 0.268 & \multicolumn{1}{l|}{20.50} & 16.61 & 0.287 & 21.58 \\
 & GradBst & \textbf{15.23} & \textbf{0.253} & \multicolumn{1}{l|}{20.24} & \textbf{15.15} & \textbf{0.259} & \multicolumn{1}{l|}{\textbf{20.00}} & \textbf{15.4} & \textbf{0.264} & \textbf{20.38} \\
 & SGDReg & 18.39 & 0.311 & \multicolumn{1}{l|}{24.05} & 18.08 & 0.310 & \multicolumn{1}{l|}{23.86} & 18.08 & 0.322 & 23.68 \\ \hline
\multirow{3}{*}{\textit{SO2}} & LR & 1.862 & 0.507 & \multicolumn{1}{l|}{\textbf{2.823}} & 2.022 & 0.547 & \multicolumn{1}{l|}{2.979} & 2.235 & 0.618 & 3.185 \\
 & GradBst & \textbf{1.607} & 0.321 & \multicolumn{1}{l|}{2.934} & \textbf{1.661} & 0.344 & \multicolumn{1}{l|}{\textbf{2.974}} & \textbf{1.678} & \textbf{0.315} & \textbf{3.09} \\
 & SGDReg & 1.784 & \textbf{0.306} & \multicolumn{1}{l|}{3.284} & 1.784 & \textbf{0.307} & \multicolumn{1}{l|}{3.325} & 1.75 & 0.321 & 3.283 \\ \hline
\end{tabular}%
}
\end{table*}

\section{Methodology}
\label{sec:method}
The collected data has been utilized to train various machine learning algorithms, which were compared to assess their effectiveness in predicting pollutant concentrations. Each model was developed to forecast a single pollutant, resulting in $|P|=5$ distinct models for each configuration.

The prediction task for each model is defined as follows: given a set of features from the different sources considered in this study, including satellite data $\Phi_S$, weather measurements $\Phi_W$, topological data $\Phi_L$, temporal features $\Phi_\mathcal{T}$, and weather forecasts $\Phi_{F}$, a separate machine learning model $g_p$ with parameters $\theta_p$ was developed for each pollutant $p \in P$. This model provides an estimate $\hat{y}_p^t$ for the given pollutant $p$ at time step $t$ on the test set, such that $\hat{y}_p^{t} = g_p( \Phi_S^{t-1}, \Phi_W^{t-1}, \Phi_L^{t-1}, ...,\Phi_S^{t-w}, \Phi_W^{t-w}, \Phi_L^{t-w}, \Phi_\mathcal{T}^{t-1}, \Phi_{F}^{t} | \theta_p)$, where $w$ indicates the specific time window applied.

Three different machine learning models were considered in this study and trained with the same set of aforementioned features: linear regression, gradient boosting regression, and stochastic gradient descent (SGD) regression.
Linear regression is a fundamental statistical method used to model the relationship between one or more independent variables and a continuous target variable. It assumes a linear relationship between the predictors and the target variable, aiming to minimize the residual sum of squares. The simplicity of the model makes it intuitive and easy to interpret, rendering it ideal for various applications.
Gradient boosting regression is a more complex machine learning model compared to linear regression. It is an ensemble learning technique that combines multiple decision trees to build a strong predictive model. It sequentially fits new models to the residuals of the previous models, effectively minimizing the loss function. In this way, the model can adapt to non-linear relationships. Gradient boosting regression is renowned for its high predictive accuracy and robustness against overfitting.
SGD regression represents a variant of linear regression that leverages an iterative approach to optimize the learning process. It is particularly advantageous when dealing with high-dimensional data and can handle large datasets with ease, offering scalability and flexibility in model training.

\section{Results}
\label{sec:results}
Due to limited data availability, the described models underwent evaluation through cross-validation, which involved partitioning the data by acquisition year. This methodology involves separating each year in the dataset and utilizing it as a validation set while training the model on the remaining years.
This process is repeated for each year, and the final performance metric is derived by averaging the values obtained across all validation sets.

To evaluate the effectiveness of the proposed models, we employed three standard performance evaluation metrics: Mean Absolute Error (MAE), Mean Absolute Percentage Error (MAPE), and Root Mean Squared Error (RMSE). 
The Mean Absolute Error (MAE) calculates the average of the absolute differences between the model predictions and the observed values. Meanwhile, the Mean Absolute Percentage Error (MAPE) computes the average of the absolute values of the percentage errors between the model predictions and the observed values, expressing the error in percentage terms relative to the actual value. Lastly, the Root Mean Squared Error (RMSE) represents the square root of the average of the squares of the differences between the model predictions and the observed values. RMSE provides an estimate of the standard deviation of the model's errors. These metrics collectively provide insights into the accuracy and performance of the machine learning models in predicting pollutant concentrations.
Furthermore, to assess the impact of the window value $w$ of past data on the results, three different values were tested: $[1, 7, 14]$, respectively corresponding to a day, a week, and two weeks of past data. All the results are summarized in Table \ref{tab:results}. 

The table highlights the superior performance of the gradient boosting model across all metrics, pollutants, and regardless of the window value $w$. However, there are a few exceptions. Specifically, for the pollutant SO2 with a window value of one or seven days, gradient boosting achieves lower MAE values, yet SGD exhibits a better MAPE, while linear regression records a superior RMSE (only observed with a one-day window). 
Moving the focus to the temporal window value used, it's evident that the model behaviour varies depending on the pollutant under consideration. For PM10, a longer window progressively leads to better results. Conversely, for PM25 and NO2, improvement is observed only when transitioning from a one-day to a seven-day window, with worse results observed with a longer window. Lastly, the most effective models trained for predicting O3 and SO2 achieve the best results with a one-day temporal window.

\section{Conclusion and future works}
\label{sec:conclusion}
The promising outcomes obtained highlight the potential of employing the proposed models as effective tools for estimating the concentrations of the five most significant pollutants within urban areas. Importantly, the utility of these models extends beyond regions with existing measurement stations. By dividing the urban area into a grid of $500$-meter cells, akin to the extent used for model training around the stations, predictions can be extrapolated for the entire urban terrain. This approach facilitates the assessment of pollutant concentration variations across different city areas.
However, opportunities for enhancement persist in several areas. Firstly, there is scope to extend the temporal horizon of predictions. While forecasting for the following day is undoubtedly valuable, broader temporal forecasts spanning multiple days or even a week ahead could significantly enhance the tool's utility. Secondly, transitioning from individual pollutant models to a unified model capable of predicting concentrations for all pollutants may yield improved performance. Despite the challenges posed by missing data resulting from variations in station configurations, such an approach could capture and exploit inter-dependencies among different pollutants. Lastly, given recent advancements in deep learning models, future approaches could explore leveraging these models to better characterize pollutant behavior within an urban context, providing increased data availability.
Nonetheless, it is essential to recognize that this work represents an initial endeavour aimed at demonstrating the feasibility of employing machine learning in this domain.


\bibliographystyle{unsrt}
\bibliography{references}

\begin{thebibliography}{10}

\bibitem{EEA2023}
European~Environment Agency.
\newblock Europe’s air quality status 2023.
\newblock \url{https://www.eea.europa.eu/publications/europes-air-quality-status-2023}, 2023.

\bibitem{LegAmb}
Legambiente.
\newblock Mal'aria di città.
\newblock \url{https://www.legambiente.it/rapporti-e-osservatori/rapporti-in-evidenza/malaria-di-citta/}, 2023.

\bibitem{ZHANG2022119347}
Bo~Zhang, Yi~Rong, Ruihan Yong, Dongming Qin, Maozhen Li, Guojian Zou, and Jianguo Pan.
\newblock Deep learning for air pollutant concentration prediction: A review.
\newblock {\em Atmospheric Environment}, 290:119347, 2022.

\bibitem{cmaq2}
Tin Thongthammachart, Shin Araki, Hikari Shimadera, Shinnosuke Eto, Tomohito Matsuo, and Akira Kondo.
\newblock An integrated model combining random forests and wrf/cmaq model for high accuracy spatiotemporal pm2.5 predictions in the kansai region of japan.
\newblock {\em Atmospheric Environment}, 262:118620, 2021.

\bibitem{NAQPMS2}
ZiFa Wang, Jie Li, Zhe Wang, WenYi Yang, Xiao Tang, BaoZhu Ge, PinZhong Yan, LiLi Zhu, XueShun Chen, HuanSheng Chen, Wei Wand, JianJun Li, Bing Liu, XiaoYan Wang, YiLin Zhao, Ning Lu, and DeBin Su.
\newblock Modeling study of regional severe hazes over mid-eastern china in january 2013 and its implications on pollution prevention and control.
\newblock {\em Science China Earth Sciences}, 57(1):3--13, Jan 2014.

\bibitem{arima1}
Lanyi Zhang, Jane Lin, Rongzu Qiu, Xisheng Hu, Huihui Zhang, Qingyao Chen, Huamei Tan, Danting Lin, and Jiankai Wang.
\newblock Trend analysis and forecast of pm2.5 in fuzhou, china using the arima model.
\newblock {\em Ecological Indicators}, 95:702--710, 2018.

\bibitem{atmos14030475}
Andrea Tateo, Vincenzo Campanaro, Nicola Amoroso, Loredana Bellantuono, Alfonso Monaco, Ester Pantaleo, Rosaria Rinaldi, and Tommaso Maggipinto.
\newblock Predicting air quality from measured and forecast meteorological data: A case study in southern italy.
\newblock {\em Atmosphere}, 14(3), 2023.

\bibitem{app10134587}
Edoardo Arnaudo, Alessandro Farasin, and Claudio Rossi.
\newblock A comparative analysis for air quality estimation from traffic and meteorological data.
\newblock {\em Applied Sciences}, 10(13), 2020.

\bibitem{MendezML}
M.~Méndez, M.G. Merayo, and M.~Núñez.
\newblock Machine learning algorithms to forecast air quality: a survey.
\newblock {\em Artificial Intelligence Review}, 2023.

\bibitem{S5-app}
Arjuman Reshi, Subbarao Pichuka, and Akshar Tripathi.
\newblock Applications of sentinel-5p tropomi satellite sensor: A review.
\newblock {\em IEEE Sensors Journal}, PP:1--1, 01 2024.

\bibitem{LI20113663}
Can Li, N.~Christina Hsu, and Si-Chee Tsay.
\newblock A study on the potential applications of satellite data in air quality monitoring and forecasting.
\newblock {\em Atmospheric Environment}, 45(22):3663--3675, 2011.

\end{thebibliography}

\end{document}